\documentclass[a4paper,10pt,oneside]{scrbook}

\usepackage{amsmath}
\usepackage{amsfonts}
\usepackage{amssymb,bm}
\usepackage{pdflscape}
\usepackage[english]{babel}
\usepackage{babelbib}
\usepackage{dsfont}
\usepackage[utf8]{inputenc}
\usepackage[pdftex]{graphicx}
\newcommand{\changefont}[3]{ 
\fontfamily{#1} \fontseries{#2} \fontshape{#3} \selectfont} 

 \normalsize
\usepackage{fancyhdr}
\fancypagestyle{plain}{%
   \fancyhead{} 
}
\begin{document}

\pagestyle{empty}
\begin{center}
\begin{tabular}{c}
\end{tabular}
\vskip 60pt
{
  \Large Taking the redpill:
  \vskip 10pt
  \textbf{Artificial Evolution in native x86 systems}
  
}
{

\vskip 40pt
}
{
  \large
  \textbf{by Sperl Thomas}\\
  \texttt{sperl.thomas@gmail.com}\\
  \vskip 10pt
  \texttt{October 2010}

\vskip 180pt
  \textbf{Abstract:}
}

\end{center}
First, three successful environments for artificial evolution in computer systems are analysed briefly. The organism in these enviroment are in a virtual machine with special \texttt{chemistries}. Two key-features are found to be very robust under mutations: Non-direct addressing and separation of instruction and argument.

In contrast, the x86 instruction set is very brittle under mutations, thus not able to achieve evolution directly. However, by making use of a special meta-language, these two key-features can be realized in a x86 system. This meta-language and its implementation is presented in chapter 2.

First experiments show very promising behaviour of the population. A statistically analyse of these population is done in chapter 3. One key-result has been found by comparison of the robustness of x86 instruction set and the meta-language: A statistical analyse of mutation densities shows that the meta-language is much more robust under mutations than the x86 instruction set.

In the end, some Open Questions are stated which should be addressed in further researches. An detailed explanation of how to run the experiment is given in the Appendix.

\tableofcontents
\pagestyle{fancy}

\chapter{Overview}
\section{History}
\subsection{CoreWorld}
Artificial evolution for self-replicating computer codes has been introduced for the first time in 1990, when Steen Rasmussen created \textit{CoreWorld}.\cite{AdamiItAL} CoreWorld is a virtual machine which can be controlled by a language called \textit{RedCode}.
This assembler-like language has a pool of ten different instructions that take two addresses as arguments. Rasmussen's idea was to introduce a random flaw of the \texttt{MOV} command, resulting of random mutations of the self-replicating codes within the environment.
The big disadvantage of RedCode was, that nearly all flaws led to a lethal mutation, hence evolution did not occure as wished.

\subsection{Tierra}

In 1992, Tom Ray found out that the problem with RedCode was due to \textit{argumented instruction set}: Independent mutations in the instruction and its arguments are unlikely to lead to a meaningful combination.\cite{RayAattsol}
Instead of direct connection between the instruction and its argument, Ray developed a pattern-based addressing mechanism: He introduced two NOP-instructions (\texttt{NOP0} and \texttt{NOP1}). These instructions do not operate themselve, but can be used as marker within the code.
A pattern-matching algorithmus would find the first appearence of a complementary marker string given (after the search-command), and returns its addresse. \\
\changefont{pcr}{m}{n}
\begin{tabbing}
    \hspace*{1cm}\=\hspace*{2cm}\= \kill

\>PUSH\_AX \>; push ax\\
\>JMP \\
\>NOP0 \\
\>NOP1 \\
\>NOP0 \>; jmp marker101 \\
\>INC\_A \>; inc ax\\
\>NOP1 \\
\>NOP0 \\
\>NOP1\>; marker101:\\
\>POP\_CX\>; pop cx\\
  \end{tabbing}

\normalfont

There are 32 instructions available in the virtual Tierra world, roughly based on assembler (\texttt{JMP, PUSH\_AX, INC\_B} and so on).
With these inventions, Ray was able to gain great results for artificial evolution (like parasitism, multi-cellularity\cite{ThearlingEPC}\cite{RayEoDMDO}, ...).

\subsection{Avida}

In 1994, Christoph Adami has developed another artificial evolution simulation, called \textit{Avida}. Beside of some different structures of the simulation, an important change has been made in the artificial chemistry: Instead of hardcoded arguments within the instructions (as in Tierra for example \texttt{PUSH\_AX}), instructions and arguments are completely separated.
The arguments are defined by NOPs (in avida there are three NOPs: \texttt{nop-A, nop-B, nop-C}) following the operation (for example, a \texttt{nop-A} following a \texttt{PUSH} pushes the AX-register to the stack). There are 24 instructions available in avida, again roughtly based on assembler (\texttt{call, return, add, sub, allocate} and so on).

\changefont{pcr}{m}{n}
\begin{tabbing}
    \hspace*{1cm}\=\hspace*{2cm}\= \kill

\>push\\
\>nop-A \>; push ax\\
\>jump-f \>\\
\>nop-A \>\\
\>nop-B \\
\>nop-B \>; jmp markerBCC\\
\>inc \\
\>nop-A \>; inc ax\\
\>nop-B \\
\>nop-C \\
\>nop-C \>; markerBCC:\\
\>pop \\
\>nop-B \>; pop bx\\
  \end{tabbing}

\normalfont

With that improvements of the virtual simulation, the researchers using avida found out amazing results, among other things about the origin of complex features in organism\cite{LenskiTeoocf}.

\section{Evolutionary Properties of different Chemistries}
\label{EvolvChem}
In 2002, an detailed analyse about different artificial chemistries has been published\cite{OfriaDoECL}. The authors compare several different instruction sets for evolutionary properties as Fitness and Robustness ($R=\frac{N}{M}$ where N is the number of non-lethal mutations and M is the number of all possible mutations).\\
The Chemistry I consists of 28 operations and has total seperated operations and arguments (same as Avida). Chemistry II has 84 unique instructions and seperated operations and arguments. The last Chemistry III has 27 instructions, but within the instructions the argument is given (i.e. \texttt{push-AX, pop-CX, ...}).
As a result, it has been found that Chemistry I is much robuster and can achieve a much higher fittness than Chemistry II, Chemistry III is the worst language for evolution.

\section{Biological Information Storage}
\label{bioinfostor}
The information in natural organism is stored in the DNA. The DNA is roughly speaking a string of nucleotides. There are four nucleotides - Adenine, Cytosine, Guanine and Thymine. Three nucleotides form a codon, and are translated by tRNA to amino acids. Amino acids are the building blocks of proteins, which are main modules of cells. \\
One can calculate that there are $N=4^3=64$ possibilities how you can sort the codons - one could code 64 different amino acids. But nature just provides 20 different amino acids, hence there is big redundancy in the translation process.

\changefont{pcr}{m}{n}
\begin{tabbing}
    \hspace*{1cm}\=\hspace*{3cm}\=\hspace*{3cm}\=\hspace*{10cm}\= \kill
    \>\textbf{Amino acid}\>\textbf{\% in human}\>\textbf{Codons}\\
    \>ALA\>6.99\>GCU, GCC, GCA, GCG\\
    \>ARG\>5.28\>CGU, CGC, CGA, CGG, AGA, AGG\\
    \>ASN\>3.92\>AAU, AAC\\
    \>ASP\>5.07\>GAU, GAC\\
    \>CYS\>2.44\>UGU, UGC\\
    \>GLU\>6.82\>GAA, GAG\\
    \>GLN\>4.47\>CAA, CAG\\
    \>GLY\>7.10\>GGU, GGC, GGA, GGG\\
    \>HIS\>2.26\>CAU, CAC\\
    \>ILE\>4.50\>AUU, AUC, AUA\\
    \>LEU\>9.56\>UUA, UUG, CUU, CUC, CUA\\
    \>LYS\>5.71\>AAA, AAG\\
    \>MET\>2.23\>AUG\\
    \>PHE\>3.84\>UUU, UUC\\
    \>PRO\>5.67\>CCU, CCC, CCA, CCG\\
    \>SER\>7.25\>UCU, UCC, UCA, UCG, AGU, AGC\\
    \>THR\>5.68\>ACU, ACA, ACC, ACG\\
    \>TRP\>1.38\>UGG\\
    \>TYR\>3.13\>UAC, UAU\\
    \>VAL\>6.35\>GUU, GUC, GUA, GUG\\
    \>\textit{STOP}\>\textit{0.24}\>\textit{UAA, UAG, UGU}\\

  \end{tabbing}
\normalfont

There is a connection between the frequency of the amino acid in the genom and the redundancy of the translation process. This mechanism protects the organism from consequences of mutation.
Imagine an Alanine (ALA) codon GCU will be mutated to GCC, this codon will still be translated to Alanine, thus there is no effect.

\chapter{Artificial Evolution in x86}

\section{Chemistry for x86}
The aim is to create an evolvable chemistry for a native x86 system. So far, all noteable attempts have been done in virtual simulated plattforms, where the creator can define the structure and the embedded instruction set.

On the other hand, the x86 chemistry has been defined long time ago and appears to be not very evolution-friendly. The instruction set is very big, the arguments and operations are directly connected, there is no instruction-end marker or constant instruction-size.
Hence, selfreplicators are very brittle in that environment, and almost all mutations are lethal.

A possibility to avoid the bad behaviour of the x86 instruction set concerning mutations is to create a (at best Turing-complete) meta-language. At execution, the meta-language has to be translated to x86 assembler instructions.

Here, a meta-language is presented with a eight bit code for each instruction, which will be translated to x86 code at execution. Obviously, this is the same procedure as in Protein biosynthesis. The eight bits coding a single instruction in the meta-language are analogs of the three codons representing one amino acid. At execution they are translated to a x86 instruction - just as tRNA transformes the codon to a amino-acid. A punch of translated x86 instructions form a specific functionality, in biology a number of amino acids form a protein (which is responsible for a certain functionality in the organism).

\section{The instruction set}

One intention was to create an instruction set which can be translated to x86 instructions in a very trivial way. This was a noteable restriction as key instructions used in Tierra and avida (like \texttt{search-f, jump-b, dividate, allocate, ...}) can not be written in a simple way in x86 assembler.

The main idea of the meta-language is to have a number of buffers which are used as arguments of all operations. Registers are not used directly as arguments for instructions, but have to be copied from/to buffers, leading to a seperation of operation and argument. The instructions have a very similar form as in avida; comparing \texttt{nopsA \& push} vs. \texttt{push \& nop-A}, or \texttt{pop \& nopdA} vs. \texttt{pop \& nop-A} or \texttt{nopsA \& inc \& nopdA} vs. \texttt{inc \& nop-A} for the meta-language and avida, respectively.

It has emerged that it is enough to use three registers (\texttt{RegA, RegB, RegD}), two buffers for calculations and operations (\texttt{BC1, BC2}) and two buffers for addressing (\texttt{BA1, BA2}).
\\
\\
\changefont{pcr}{m}{n}
\begin{tabbing}
    \hspace*{2cm}\=\hspace*{2cm}\=\hspace*{2cm}\=\hspace*{3.5cm}\= \kill
    \>\underline{Buffer instructions (16)}\\
\\
    nopsA \>\>BC1=RegA\>\>mov ebx, eax\\
    nopsB \>\>BC1=RegB\>\>mov ebx, ebp\\
    nopsD \>\>BC1=RegD\>\>mov ebx, edx\\
\\
    nopdA \>\>RegA=BC1\>\>mov eax, ebx\\
    nopdB \>\>RegB=BC1\>\>mov ebp, ebx\\
    nopdD \>\>RegD=BC1\>\>mov edx, ebx\\
\\
    saveWrtOff \>\>BA1=BC1 \>\>mov edi, ebx\\
    saveJmpOff \>\>BA2=BC1 \>\>mov esi, ebx\\
\\
    writeByte  \>\>byte[BA1]=(BC1\&\&0xFF)\>\>mov byte[edi], bl\\
    writeDWord \>\>dword[BA1]=BC1\>\>mov dword[edi], ebx\\
\\
    save \>\>BC2=BC1\>\>mov ecx, ebx\\
    addsaved \>\>BC1+=BC2 \>\>add ebx, ecx\\
    subsaved \>\>BC1-=BC2\>\>sub ebx, ecx\\    
\\ 
    getDO \>\>BC1=DataOffset\>\>mov ebx, DataOffset\\
    getdata \>\>BC1=dword[BC1]\>\>mov ebx, dword[ebx]\\
\\
    getEIP\>\>BC1=instruction pointer\>\>call gEIP; gEIP: pop ebx\\

\\
\\
    \>\underline{Operations (10+8)}\\
\\    
zer0\>\>BC1=0\>\>mov ebx, 0x0\\
push\>\>push BC1\>\>push ebx\\
pop\>\>pop BC1\>\>pop ebx\\
mul\>\>RegA*=BC1\>\>mul ebx\\
div\>\>RegA/=BC1\>\>div ebx\\
shl\>\>BC1 << (BC2\&\&0xFF)\>\>shl ebx, cl\\
shr\>\>BC1 >> (BC2\&\&0xFF)\>\>shr ebx, cl\\
and\>\>BC1=BC1\&\&BC2\>\>and ebx, ecx\\
xor\>\>BC1=BC1 xor BC2\>\>xor ebx, ecx\\
add0001\>\>BC1+=0x1\>\>add ebx, 0x0001\\
add0004\>\>BC1+=0x4\>\>add ebx, 0x0004\\
add0010\>\>BC1+=0x10\>\>add ebx, 0x0010\\
add0040\>\>BC1+=0x40\>\>add ebx, 0x0040\\
add0100\>\>BC1+=0x100\>\>add ebx, 0x0100\\
add0400\>\>BC1+=0x400\>\>add ebx, 0x0400\\
add1000\>\>BC1+=0x1000\>\>add ebx, 0x1000\\
add4000\>\>BC1+=0x4000\>\>add ebx, 0x4000\\
sub0001\>\>BC1-=1\>\>sub ebx, 0x0001\\\\
\\
    \>\underline{Jumps (4)}\\
\\
JnzUp\>\>jz over \&\& jmp esi \&\& over:\\
JnzDown\>\>jnz down (\&\& times 32: nop) \&\& down:\\
JzDown\>\>jz  down (\&\& times 32: nop) \&\& down:\\
ret\>\>ret\\

\\
\\
    \>\underline{API calls (11) - Windows based}\\
\\
CallAPIGetTickCounter \>\>\>stdcall [GetTickCount]\\
CallAPIGetCommandLine \>\>\>stdcall [GetCommandLine]\\
CallAPICopyFile \>\>\>stdcall [CopyFile]\\
CallAPICreateFile \>\>\>stdcall [CreateFile]\\
CallAPIGetFileSize \>\>\>stdcall [GetFileSize]\\
CallAPICreateFileMapping \>\>\>stdcall [CreateFileMapping]\\
CallAPIMapViewOfFile \>\>\>stdcall [MapViewOfFile]\\
CallAPICreateProcess \>\>\>stdcall [CreateProcess]\\
CallAPIUnMapViewOfFile \>\>\>stdcall [UnMapViewOfFile]\\
CallAPICloseHandle \>\>\>stdcall [CloseHandle]\\
CallAPISleep \>\>\>stdcall [Sleep]\\

  \end{tabbing}
\normalfont
There are 30+8 unique commands (the eight \texttt{addNNNN} and \texttt{sub0001} could be reduced to one single command, but this would make the code very big) and 11 API calls - giving 49 instructions. For translation, a command is identified by 8bits. Therefore there are $N=2^8=256$ possible combinations, thus there is a big redundancy within the translation of commands to x86 code - just as in natural organism.
This gives the code a big freedom in protecting itself against harmful effects of mutations.

\subsection{An example: Linear congruential generator}
The following code creates a new random number (Linear congruential generator) via 
\begin{equation}
x_{n+1}= (a x_n + c) \textnormal{ mod } m
\nonumber
\end{equation}
with $a=1103515245$, $c=12345$ and $m=2^{32}$ (these are the numbers used by GCC).\\

\vspace{30pt}

\changefont{pcr}{m}{n}
\begin{tabbing}
    \hspace*{1cm}\=\hspace*{4cm}\=\hspace*{2cm}\= \kill

.data\\
DataOffset:\\
\>SomeData dd 0x0\\
\>RandomNumber dd 0x0\\
\\
.code\\
macro addnumber arg \{ ... \}\\
\>\>; Creates the correct addNNNN combination\\
\\
\>getDO\\
\>add0004\\
\>getdata\>; mov ebx, dword[RandomNumber]\\
\>nopdA\>; eax=dword[RandomNumber]\\
\\
\>zer0\\
\>addnumber 1103515245\>\>; mov ebx, 1103515245\\
\>mul\>; mul ebx\\
\\
\>zer0\\
\>addnumber 12345\\
\>save\>; mov ecx, ebx\\
\\
\>nopsA\\
\>addsaved\\
\>nopdB\>; mov ebp, (1103515245*[RandomNumber]+12345)\\
\>\>; ebp=new random number
\\
\>getDO\\
\>add0004\\
\>saveWrtOff\>; mov edi, RandomNumber\\
\\
\>nopsB\\
\>writeDWord\>; mov dword[RandomNumber], ebp\\
\>\>; Save new random number\\
.end code
  \end{tabbing}
\normalfont
\section{Translation of meta-language}
As the instruction set has been created to construct a trivial translator, the translator can be written as a single loop. A meta-language instruction is one byte, the corresponding x86 instruction has 8 bytes (for 256 instructions, this gives a $8*256=2.120$ Byte long translation table).

The Translator picks one 8bit codon, searchs the corresponding x86 instruction and writes that x86 instruction to the memory. At the end of all codons, it executes the memory.

\changefont{pcr}{m}{n}
\begin{tabbing}
    \hspace*{0cm}\=\hspace*{3cm}\=\hspace*{1cm}\=\hspace*{5cm}\= \kill
        \>invoke  \>VirtualAlloc, 0x0, 0x10000, 0x1000, 0x4 \\
        \>mov     \>[Place4Life], eax\>\>; 64 KB RAM\\
        \>mov     \>edx, 0x0                                \>\>; EDX will be used as the\\
        \>\>                                                \>\>; counter of this loop\\
        
        \>WriteMoreToMemory:\\
        \>\>    mov     \>ebx, 0x0                        \>; EBX=0\\
        \>\>    mov     \>bl, byte[edx+StAmino]           \>; BL=NUMBER OF AMINO ACID\\

       \>\>     shl    \>ebx, 3                         \>; EBX*=8;\\
       \>\>     mov    \>esi, StartAlphabeth            \>; Alphabeth offset\\
       \>\>     add   \>esi, ebx                        \>; offset of the current amino acid\\

       \>\>         mov     \>ebx, edx                        \>; current number of amino acid\\
       \>\>         shl     \>ebx, 3                          \>; lenght of amino acids\\
       \>\>         mov     \>edi, [Place4Life]               \>; Memory address\\
       \>\>         add     \>edi, ebx                        \>; Offset of current memory\\

       \>\>         mov     \>ecx, 8                          \>; ECX=8\\
       \>\>         rep     \>movsb                           \>; Write ECX bytes from ESI to EDI\\
                                                        \>\>\>\>; Write 8 bytes from Alphabeth\\
                                                        \>\>\>\>; to Memory\\
       \>\>         inc     \>edx                             \>; Increase EDX\\
        \>cmp\>     edx, (EndAmino-StAmino)\\
        \>jne\>     WriteMoreToMemory\\
\\
        \>call\>    [Place4Life]\>\>; Run organism!\\
        
  \end{tabbing}
\normalfont
The Translation Table/Alphabeth has the following form:
\changefont{pcr}{m}{n}
\begin{tabbing}
    \hspace*{3cm}\=\hspace*{1cm}\=\hspace*{5cm}\= \kill
; 0001 1000 - 24:\\
\_getEIP EQU 24\\
ACommand24:\\
\>        call gEIP\\
\>        gEIP:\\
\>        pop ebx\\
ECommand24:\\
times (8-ECommand24+ACommand24): nop\\
\\
\\
; 0001 1001 - 25:\\
\_JnzUp EQU 25\\
ACommand25:\\
\>        jz over\\
\>        jmp esi\\
\>        over:\\
ECommand25:\\
times (8-ECommand25+ACommand25): nop
  \end{tabbing}
\normalfont
\chapter{Experiments}
\label{experiment}
\textit{To achieve evolution it is necessary to have replication, mutation and selection.}
\section{Overview}
An ancestor organism has been written, which is able \textbf{replicate} itself. It copies itself in the current directory to a random named file and execute its offspring.\vspace{0.1cm} 

The \textbf{mutation}-algorithm is written within the code (not given by the plattform as it is possible in Tierra or avida). With a certain probability a random bit within a special interval of the new file flips. Each organism can create five offspring, each with a different intervall and probability of mutation.

For finding an adequate mutation probability, one can calculate the probability $P$ that at least one bit-flip occures giving a $N$ byte interval and a probability $p_{bit}$ for a single bit to flip:

\begin{equation}
P(N,p_{bit})=\sum_{n=0}^{N-1} p_{bit} (1- p_{bit})^n = 1 - (1 - p_{bit})^n
\nonumber
\end{equation} 

\changefont{pcr}{m}{n}
\begin{tabbing}
    \hspace*{2cm}\=\hspace*{0.5cm}\=\hspace*{4cm}\=\hspace*{1.5cm}\=\hspace*{1.5cm}\= \kill
    \>\>\textbf{Interval}\>\textbf{N}\>\textbf{P}\>\textbf{$p_{bit}$}\\
    \>1\>Code\>2100\>0.9\>$\frac{1}{900}$\\
    \>2\>Code+Alphabeth\>4200\>0.9\>$\frac{1}{1800}$\\
    \>3\>whole file\>6150\>0.9\>$\frac{1}{2666}$\\
    \>4\>Code\>2100\>0.75\>$\frac{1}{1500}$\\
    \>5\>Code\>2100\>0.68\>$\frac{1}{1820}$\\

  \end{tabbing}
\normalfont

The second offspring has also the opportunity to change the alphabeth. This could lead to redundancy in the alphabeth to avoid negative effects of mutations (as used in nature - descriped in \ref{bioinfostor}). The mutations in the third offspring can access the whole file.\vspace{0.1cm} 

Natural \textbf{selection} is not very strong in this experiment, CPU speed and harddisk space is limited. Thus, most non-lethal mutations are neutral and disribute randomly within the population. This can be used very easy to understand the relationship of the population: The smaller their \textit{Hamming distance}, the nearer their relationship.

The Hamming distance $\Delta(x, y)$ is defined as

\begin{equation}
\Delta(x, y):=\sum_{x_i \not= y_i} 1,\quad i=1,...,n
\nonumber
\end{equation}
Beside of natural selection, their could be artificial selection. Some artificial selection has been used to prevent some wired behaviour of the populations.

The experiments have been done on a native WindowsXP. For stabilization, several small C++ guard programs have been developed which searches and closes endless-loops, closes error messages and dead processes (processes that live longer than a certain time). 

\section{First attempts}
\label{MultInst}
The first attempt has shown some unexpected behaviour.\vspace{0.5cm}\\
\textbf{Multiple instances of same file:} Already after a few dozen of generations, the process list started to fill with multiple instances of the same file. An analyse of the file shows that this happens due to a mutation corrupting the random name engine. The random name engine always generates the same filename (for instance \texttt{aaaaaaaa.exe}). After the mutation process (which has no effect as the file has write protection due to execution) the new/old file is executed again.

To prevent this unnatural behaviour, it has become necessary to include an artificial selection to the system. The new C++ file scans the process list for multiple file instances, and closes them.

It is interesting to see that this is a real selection, not a restriction to the system. Mutation still can create such effects, but with that additional guard file, they have negative consequences for the file (the will be deleted immediatly) and therefore will not spread within the population.\vspace{0.5cm}\\
\textbf{Avoid mutations:} The first long-time experiment appeared to be very promising. The guard files closed error messages, endless loops, dead processes and multiple instances of the same file. After some hundred generations the experiment has been stopped and the files have been analysed. Surprisingly, all files had the exact same bit-code, they were all clones.

There has been a mutation in the alphabeth, changing the \texttt{xor} instruction. This instruction is responsible for changing a bit at the mutation-process. If the mutation does not work anymore, no files will change anymore.

For the organism, this is a big advantage. All offspring will survive as no more mutation happens. Other organism often create corrupt offspring, hence spread slower. After a while, the whole system is dominated by unmutable organism.

In nature, organism also created very complex systems to prevent mutation or mutational effects. DNA repairing or amino acid redundancy are just two examples.

Even this discoverment is very interesting and has a great analogon in nature, it prevents from further discoverments in this artificial system. Therefore another guard file has been developed, which scans the running files for clones and deletes them.

It's not natural to prohibit clones at all, thus a adequate probability should be found. If there are 42 clones in the process list, they should be detected by a probability of 51\% in one guard file cycle. This gives a probability of $P=\frac{1}{59}$ that a running file will be checked whether it has clones. A controlled file will be compared to all other running files, all clones will be deleted.

\section{Statistical analyse of experiment}
Afer installing the new guard file, a further experiment has been run. This first "long-term" experiment can be analysed statistically by comparing the density and type of mutation from the ancestor file and the latest population. Unforunatley it is very hard to determinate the number of generations in the population; by comparing mutations in the oldest population with the primary ancestor and using the mutation probability, one could speculate that there have been 400-600 generations.\vspace{0.5cm}\\
\textbf{Number of mutations - ancestor vs. successor:} 
A number of 100 successor have been randomly picked and compared with the ancestor. One can calucate the average number of mutation during the lifetime of the experiment, and its standard deviation:\\

\begin{equation}
\bar{X}= \frac{1}{n} \sum_{i=1}^n{X_i} = 192.02 \textnormal{ Mutations}
\nonumber
\end{equation}
\begin{equation}
\sigma= \sqrt{\frac{1}{n-1} \sum_{i=1}^n{(X_i-\bar{X})^2}} = 4.59
\nonumber
\end{equation}
The standard deviation gives an unexpected small value, which means that the number of mutations is quite constant over the lifetime of the population.\vspace{0.5cm}\\
\textbf{Relations between individua:}
One can analyse the relation beween the individua by calculating their Hamming distance (the number of differences in their bytecode). A number of six files have been selected randomly and analysed.

\changefont{pcr}{m}{n}
\begin{tabbing}
    \hspace*{4.5cm}\= \kill
\>ancestor.exe - a.exe: 195\\
\>ancestor.exe - b.exe: 195\\
\>ancestor.exe - c.exe: 184\\
\>ancestor.exe - d.exe: 192\\
\>ancestor.exe - e.exe: 194\\
\>ancestor.exe - f.exe: 200\\

  \end{tabbing}
\normalfont

\changefont{pcr}{m}{n}
\begin{tabbing}
    \hspace*{5.5cm}\=\hspace*{5.5cm}\= \kill
a.exe - b.exe: 2\>c.exe - a.exe: 75\>e.exe - a.exe: 9\\
a.exe - c.exe: 75\>c.exe - b.exe: 75\>e.exe - b.exe: 9\\
a.exe - d.exe: 20\>c.exe - d.exe: 73\>e.exe - c.exe: 74\\
a.exe - e.exe: 9\>c.exe - e.exe: 74\>e.exe - d.exe: 19\\
a.exe - f.exe: 28\>c.exe - f.exe: 81\>e.exe - f.exe: 27\\
\\
b.exe - a.exe: 2\>d.exe - a.exe: 20\>f.exe - a.exe: 28\\
b.exe - c.exe: 75\>d.exe - b.exe: 20\>f.exe - b.exe: 28\\
b.exe - d.exe: 20\>d.exe - c.exe: 73\>f.exe - c.exe: 81\\
b.exe - e.exe: 9\>d.exe - e.exe: 19\>f.exe - d.exe: 16\\
b.exe - f.exe: 28\>d.exe - f.exe: 16\>f.exe - e.exe: 27\\
  \end{tabbing}
\normalfont
While \texttt{a.exe, b.exe} and \texttt{e.exe} are near related, \texttt{c.exe} is far away from all other files. \texttt{d.exe} and \texttt{f.exe} are medium related. Interestingly, while \texttt{c.exe} has the biggest distance to all other successors, it has the smallest distance to the ancestor.\vspace{0.5cm}\\
\textbf{Distribution of mutations:}
It is interesting to see which mutations are rare and which are widely spread within the population. There are 153 mutations which appeare in every single file, 32 mutations appearing in 84 files and so on. Many mutations are located in unused areas of the file, for instance in the Win32 .EXE padding bytes or in the unused part of the alphabeth. A list of mutations of the active-code (whether the used part of the alphabeth or the meta-language code) and its appearence in the population is given here. 

\changefont{pcr}{m}{n}
\begin{tabbing}
    \hspace*{3cm}\=\hspace*{3cm}\=\hspace*{3cm}\=\hspace*{3cm}\= \kill
527: 100\>e52: 100\>12af: 100\>147f: 100\>e13: 32\\
551: 100\>e9a: 100\>12b1: 100\>1498: 100\>1298: 23\\
56c: 100\>eac: 100\>12d9: 100\>14a5: 100\>f0a: 17\\
5af: 100\>f04: 100\>130e: 100\>14b3: 100\>4c3: 16\\
5ed: 100\>f34: 100\>131f: 100\>14b9: 100\>558: 16\\
61a: 100\>fbb: 100\>1327: 100\>14c3: 100\>58a: 16\\
625: 100\>fed: 100\>1328: 100\>5b9: 84\>60d: 16\\
c74: 100\>1090: 100\>1333: 100\>d86: 84\>ca9: 16\\
c7b: 100\>10b8: 100\>1343: 100\>e43: 84\>cac: 16\\
c98: 100\>10c4: 100\>135c: 100\>1037: 84\>d83: 16\\
c9b: 100\>1119: 100\>1373: 100\>106b: 84\>df4: 16\\
ca1: 100\>1121: 100\>138d: 100\>109c: 84\>f5a: 16\\
ca4: 100\>1126: 100\>139c: 100\>1148: 84\>105c: 16\\
d02: 100\>118f: 100\>13b3: 100\>127d: 84\>1085: 16\\
d4f: 100\>1194: 100\>13d5: 100\>12a8: 84\>10ef: 16\\
d5d: 100\>11a2: 100\>13eb: 100\>130f: 84\>12d1: 16\\
d7a: 100\>11a9: 100\>13fd: 100\>1388: 84\>12ee: 16\\
d7d: 100\>11b1: 100\>1430: 100\>1392: 84\>1323: 16\\
e3b: 100\>124d: 100\>144c: 100\>13e4: 84\>1353: 16\\
e49: 100\>1265: 100\>1459: 100\>c9d: 32\>139a: 16\\
\end{tabbing}
\normalfont
A full analyse of these mutations would be worthwhile, but has not be done in this primary analyse due to its great effort. However, to understand this system and its prospects better, detailed code analyse will be unavoidable.

Nevertheless, examples of two mutations can be given.\\

First one is \textbf{Byte 0x527}: This is within the alphabeth, defining the behaviour for the \texttt{JnzUp} instruction. A bit-flip caused following variation:

\changefont{pcr}{m}{n}
\begin{tabbing}
    \hspace*{1cm}\=\hspace*{3cm}\=\hspace*{1cm}\= \kill
\>jz over\>\>jz over\\
\>jmp esi\>$	\rightarrow$\>jmp esi\\
\>over:\>\>nop\\
\>\>\>over:\\
  \end{tabbing}
\normalfont
This has no effect in the behaviour, but effects just the byte-code - a neutral mutation.\\

The second example is the mutation in \textbf{Byte 0xC7B}, which is within the meta-language code. The unmutated version is the instruction \texttt{add0001}, the mutated one represents \texttt{add0004}. This is part of the \texttt{addnumber 26} instruction, which is used as modulo number for the random name generator.

Due to this mutation, the genom not just picks letters from $a-z$ for its offpring's filename, but also the next three in the ASCII list, \{, $\vert$ and \}. Thus, filenames can also contain these three characters. This mutation has an effect of the behaviour, but still seems to be a neutral mutation.  

\section{Comparing Robustness with x86 instruction set}
In 2005 a program called \texttt{Gloeobacter violaceus} has been developed, that uses artificial mutations in the x86 instruction set, without making use of a meta-language.\cite{spthgv} That program also replicates in the current directory, and is subjected by point mutations, and rarely by inseration, deletion and dublication. Due to the brittleness of the x86 instruction set, that attempt was not very fruitful. Still it gives a good possibility of comparison.\\

Both systems have changed to the same initial situation: Point mutations occure in the whole file with same probability. After several hundreds of generations, all non-minor mutations (occure in more than 50 different files) of 2.500 files have been analysed. The mutations have been classified by their position: x86-code mutations, mutation in some padding region or in the meta-language code.\\

\textit{Through this classification we find out whether the new meta-language concept is more robust than the x86 instruction set.}\\

We define the mutation density of a specfic region in the code by 
\begin{equation}
\rho_{mut}(\textnormal{Region})=\frac{\textnormal{mutations in region}}{\textnormal{size of region}}
\nonumber
\end{equation}

\textbf{Meta-Language concept}:
\begin{align*}
\rho_{mut}({\textnormal{whole code})}=\frac{291}{6144}=0.047\nonumber\\
\rho_{mut}({\textnormal{padding})}=\frac{151}{2427}=0.062\nonumber\\
\rho_{mut}({\textnormal{meta-code})}=\frac{81}{2084}=0.039\nonumber\\
\rho^{*}_{mut}({\textnormal{x86})}=\frac{14}{576}=0.024\nonumber\\
\nonumber
\end{align*}

The $\rho_{mut}({\textnormal{x86})}$ combines the very small translator code and the alphabeth, but as the alphabeth is no real x86 code, comparing this advisable. If that problem would be neglected, one would see that the meta-language is more robust under mutations as the x86 code. For a fair comparison \texttt{Gloeobacter violaceus} can be used.

\textbf{Gloeobacter violaceus}:
\begin{align}
\rho_{mut}({\textnormal{whole code})}=\frac{351}{3584}=0.098\nonumber\\
\rho_{mut}({\textnormal{padding})}=\frac{284}{2229}=0.127\nonumber\\
\rho_{mut}({\textnormal{x86})}=\frac{10}{683}=0.015\nonumber\\
\nonumber
\end{align}

Mutations in the padding bytes do not corrupt the organism, thus it is the initial mutation density. A comparison between $\rho_{mut}({\textnormal{padding})}$ and $\rho_{mut}({\textnormal{Region})}$ gives the percentage of non-lethal mutations in that region, therefore gives the robustness R of that region. 

\begin{equation}
\textnormal{Robustness}(\textnormal{Region}):=\frac{\rho_{mut}({\textnormal{Region})}}{\rho_{mut}({\textnormal{padding})}}\nonumber
\end{equation}
The interesting comparison is between the x86 region and the meta-language region. 

\begin{equation}
\textnormal{Robustness}(\textnormal{x86})=\frac{\rho_{mut}({\textnormal{x86})}}{\rho_{mut}({\textnormal{padding})}}=\frac{0.015}{0.127}=0.115\nonumber
\end{equation}

\begin{equation}
\textnormal{Robustness}(\textnormal{meta-code})=\frac{\rho_{mut}({\textnormal{meta-code})}}{\rho_{mut}({\textnormal{padding})}}=\frac{0.039}{0.092}=0.424\nonumber
\end{equation}
\\Even this analysis is based on low statistics, it already indicates a great result:\vspace{0.2cm}\\
\textbf{This new meta-language concept for x86 systems is much more robust than the original x86 instruction set.}

\chapter{Outlook}
\section{Open questions}
\textbf{Development of new functionality:} The most important question is whether an artificial organism with this meta-language in a x86 system can develope new functionalities.

In a long-term evolution experiment by Richard Lenski, they discovered that simple E.coli was able to make a major evolutionary step and suddenly acquired the ability to metabolise citrate.\cite{LenskiEColi} This happened after 31.500 generations, approximativly after 20 years.

The generation time of artificial organism are of many orders of magnitude smaller, therefore beneficial mutation such as development of new functionality may occure within days or a few weeks.
The question that remains is whether this meta-language concept is the right environment or not.\vspace{0.1cm}\\
\textbf{Other types of mutations:} Point mutation is one important type of mutation, but not the only one. In DNA, there is also Deletion, Duplication, Inseration, Translocation, Inversion. Especially inseration of code and deletion of code is proved to be important in artificial evolution too.\cite{Lenski} The question is how one can create such a type of mutation without file structure errors occuring after every single mutation.

One possibility would be to more the n last byte of the meta-language code forwards (deletion) or backwards (inseration), filling the gap with \texttt{NOPs}. However, how could you find out where the end of the meta-language code is without some complex (and thus brittle) functions?\vspace{0.1cm}\\
\textbf{Behaviour of Hamming distance:} How is the time evolution of the average Hamming distance between a population and the primary ancestor? Does it have a constant slope or is it rather like a logarithm? How is the behaviour of the Hamming distance when taking into account other types of mutations (as descriped above)? Large-scale experiments are needed to answer that questions propriatly.
\vspace{0.1cm}\\
\textbf{APIs:} This is an operation system specific problem, and can not be solved for any OS at once. For Windows, the current system of calling APIs is not very natural. It is a call to a specific addresse of a library, needing the right numbers of arguments on the stack and the API and library defined in the file structure. Hence, API calls are not (very) evolvable in this meta-language, restricting the ability to use new APIs by mutations.

One possible improvement could be the usage of \texttt{LoadLibraryA} and \texttt{GetProcAddress}, which loads the APIs from the kernel autonomous. This technique would not need the APIs and libraries saved within the file structure, and could make it possible to discover new functionalities. Unfortunately, this requires complex functions, which may be very brittle and unflexible. 

Still it needs more thoughts to find an adequate solution to this problem.

\section{Computer malware}
This technique could be used in autonomous spreading computer programs as computer viruses or worms. This has been discussed in a very interesting paper by Iliopoulos, Adami and Sz\"or in 2008.\cite{szor}

Their main results are: The x86 instruction set does not allow enough neural mutations, thus it is impossible to develope new functionalities; a 'evolvable' language or a meta-language would be needed. Further, together with smaller generation times, the selective pressure and the mutation rate would be higher, speeding up evolution. Conclusion is, that it is currently unclear what would be a defence against such viruses. 
 
In contrast to the experiment explained in chapter \ref{experiment} - where natural selection was nearly absent, computer malware are continuously under selective pressure due to antivirus scanners. This is the same situation as in biological organism, where parasits are always attacked by the immune system and antibiotics.

Theoretically, computer malware could also find new ways to exploit software or different OS APIs for spreading. This is not as unlikely as it seems in the first moment. Experiments with artificial and natural evolution have shown that complex features could evolve in acceptable time.\cite{LenskiTeoocf}\cite{Lenski}

\section{Conclusion}
An artificial 'evolvable' meta-language for x86 Systems has been created using the main ideas of \texttt{Tierra} and \texttt{avida}: Separation of operations and arguments, and not using direct addressing.
The experiments have been very promising, showing that the robustness of the new meta-language is approximately four times higher as for usual x86 instructions.
Several open questions are given in the end, which should motivate for further research.\vspace{0.5cm}\\
Howsoever, the most important step has been done:\\ \textbf{The artificial organism are not trapped in virtual systems anymore, they can finally move freely - they took the redpill...}

\begin{appendix}
\chapter{Appendix}
This artificial evolution system can be started on every common Windows Operation System. Even it is a chaotic process, due to the guard files the process can run for hours without a breakdown of the system.
\section{The package}
\textbf{The package:}\\
\texttt{\underline{binary\textbackslash run0ndgens.bat}:} This script starts all guard files, then starts the 0th generation. Adjust the hardcoded path in the file to the directory of the guard files. This file has to be at H:, you can use \texttt{subst} for that.\vspace{0.1cm}\\
\texttt{\underline{binary\textbackslash NewArt.exe}:} This is the 0th generation. It has to be started with the shell (or a .bat file) - not via double-click. It is highly recommented to not run the file without the guard files. This file has to be as H: aswell.\vspace{0.3cm}\\
\texttt{\underline{ProcessWatcher\textbackslash *.*}:} This directory contains the binary and source of all guard files.\vspace{0.1cm}\\
\texttt{\underline{ProcessWatcher\textbackslash CopyPopulation.cpp}:} This file copies every 3 minutes 10\% of the population to a specific path given in the source. This path has to be adjusted before usage.\vspace{0.1cm}\\
\texttt{\underline{ProcessWatcher\textbackslash Dead.cpp}:} This program can be used to manually stop all organism. You can enter a probability of how many organism should be survive. For instance, if you enter 10, 90\% of the population will be terminated - 10\% survive.\vspace{0.1cm}\\ 
\texttt{\underline{ProcessWatcher\textbackslash DoubleProcess.cpp}:} This program searchs and destroyes multiple instances of the same file. See Chapter \ref{MultInst} for more information.\vspace{0.1cm}\\
\texttt{\underline{ProcessWatcher\textbackslash EndLessLoops.cpp}:} This guard file searchs for endless loops in the memory and terminates them.\vspace{0.1cm}\\
\texttt{\underline{ProcessWatcher\textbackslash JustMutation.cpp}:} Also descriped in chapter \ref{MultInst}, this program searchs and terminates clones in the process.\vspace{0.1cm}\\
\texttt{\underline{ProcessWatcher\textbackslash Kill2MuchProcess.cpp}:} This guard is very important for stability of the operation system while running the experiment. If there are more than 350 processes running, it terminates 75\% of them.\vspace{0.1cm}\\
\texttt{\underline{ProcessWatcher\textbackslash RemoveCorpus.cpp}:} As space is restriced, this guard deletes files that are older than a 30sec.\vspace{0.1cm}\\
\texttt{\underline{ProcessWatcher\textbackslash SearchAndDestroy.cpp}:} This program removes error messages (by clicking at "OK"), terminates error-processes (as dwwin.exe or drwtsn32.exe), and it terminates dead processes (processes that are older than 100sec).\vspace{0.1cm}\\ 
\texttt{\underline{ProcessWatcher\textbackslash malformed\_PEn.exe}:} These are two malformed .EXE files, which will be called by \texttt{SearchAndDestroy.exe} at the start to find the "OK"-Button. They have to be in the directory of the guard files.\vspace{0.3cm}\\  
\texttt{\underline{Analyse\textbackslash SingleFileAnalyse}:} This directory contains an analyse file, that compaires the bytecode of two genotypes. Copy the file to the directory, change the name in the source and execute it.\vspace{0.1cm}\\ 
\texttt{\underline{Analyse\textbackslash Relation}:} This file compaires gives you the Hamming distance of all .exe files. \vspace{0.1cm}\\
\texttt{\underline{Analyse\textbackslash MutationDistribution}:} With this file you can get a distribution of all mutations compaired with NewArt.exe.\vspace{0.3cm}\\

\section{Running the experiment}
Copy the \texttt{run0ndgens.bat} and \texttt{NewArt.exe} to \texttt{H:\textbackslash }. Adjust the path in \texttt{CopyPopulation.cpp} to the backup directory (and compile it) and in \texttt{run0ndgens.bat} to the directory of the guard files.

Now you can start \texttt{run0ndgens.bat}, move over the two error-messages (dont click them, this will be done by \texttt{SearchAndDestroy.exe}). Then you are ready and can press a key in the \texttt{run0ndgens.bat}, which will start 10 instances of \texttt{NewArt.exe}.

\begin{center}
\begin{tabular}{c}
  \includegraphics[scale=0.22]{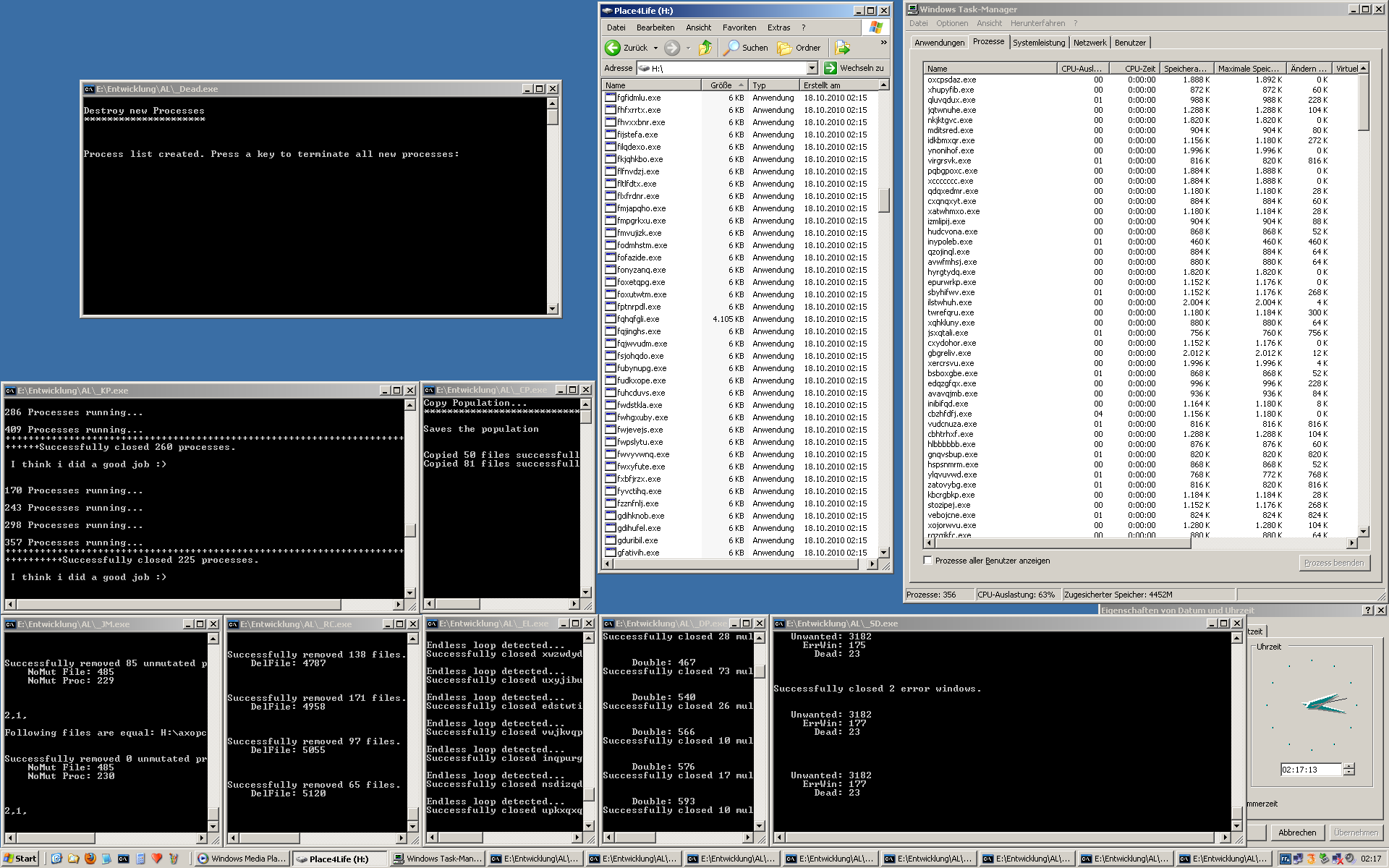} \\
  \textbf{Running Experiment:} This is how the experiment should look like\\
  \end{tabular}
\end{center}

\end{appendix}

\end{document}